\documentclass[a4paper,journal,12pt,draftcls,onecolumn]{IEEEtran}%
\usepackage{amsfonts}
\usepackage{amsmath}
\usepackage{amssymb}
\usepackage{graphicx}%
\providecommand{\U}[1]{\protect\rule{.1in}{.1in}}

\begin{document}

\title{Reliable Flight Control: Gravity-Compensation-First Principle}
\author{Quan~Quan\thanks{Q. Quan is with the School of Automation Science and
Electrical Engineering, Beihang University, Beijing 100191, China (e-mail:
qq\_buaa@buaa.edu.cn (Q. Quan)).}}
\maketitle

\begin{abstract}
Safety is always the priority in aviation. However, current state-of-the-art
passive fault-tolerant control is too conservative to use; current
state-of-the-art active fault-tolerant control requires time to perform fault
detection and diagnosis, and control switching. But it may be later to recover
impaired aircraft. Most designs depend on failures determined as a priori and
cannot deal with fault, causing the original system's state to be
uncontrollable. However, experienced human pilots can save a serve impaired
aircraft as far as they can. Motivated by this, this paper develops a
principle to try to explain human pilot behavior behind, coined the
\emph{gravity-compensation-first principle.} This further supports reliable
flight control for aircraft such as quadcopters and tail-sitter unmanned
aerial vehicles.

\end{abstract}

\section{Introduction}

Considerable attention has recently been gained to Fault-tolerant control
(FTC) in recent years because of its important role in maintaining the
systems' safety via configured redundancy \cite{Yu(2015)}. FTC is roughly
classified into passive FTC and active FTC. The control structure of a passive
FTC system will not be changed in both normal and abnormal conditions without
performing real-time management of redundancies. This is because the system
redundancy has already been integrated into the controller design. In an
active FTC scheme, fault detection and diagnosis (FDD) has to be performed via
the information from the monitoring system, based on which control algorithms
as well as manage redundancies are reconfigured to make the system safe in
abnormal conditions. From above, the active FTCs have three main steps: (1)
FDD; (2) reconfigurable control; and (3) integration of FDD and reconfigurable
control. FDD plays an important and essential role, but it imposes a severe
burden on computation, and its delay can adversely affect system safety.
Compared with the active FTC, the fault-tolerant ability of the passive FTC is
more conservative. Still, its inherent feature of no FDD and controller
switching is significant for flight control systems, usually having a minimal
amount of time to be recovered.

Most FTC approaches focus on modifying the inner loop of a flight control
system \cite{Steinberg(2015)}. However, an impaired aircraft may have
significant restrictions on both its maneuvering capability and the flight
envelope or lose its original controllability through which it can be safely
controlled. However, the human pilot, who has sometimes developed very
innovative strategies to control impaired aircraft, can deal with these
issues. First, let us study an accident shown in Fig.1, where a fixed-wing
aircraft lost its right-wing in the air at 16s in the video. Since then, the
aircraft cannot perform its normal flight task anymore. Instead, the aircraft
dropped height and started rolling. At 35s, the nose of the aircraft pointed
upward, and its rotation was stopped. From then on, the nose of the aircraft
always pointed upward until it approached the ground. At 48s, the aircraft
changed its attitude by its left aileron to land on the ground with its
wheels. Otherwise, it will sit on the ground with its tail.

\begin{figure}[h]
\centering \includegraphics[scale=0.7]{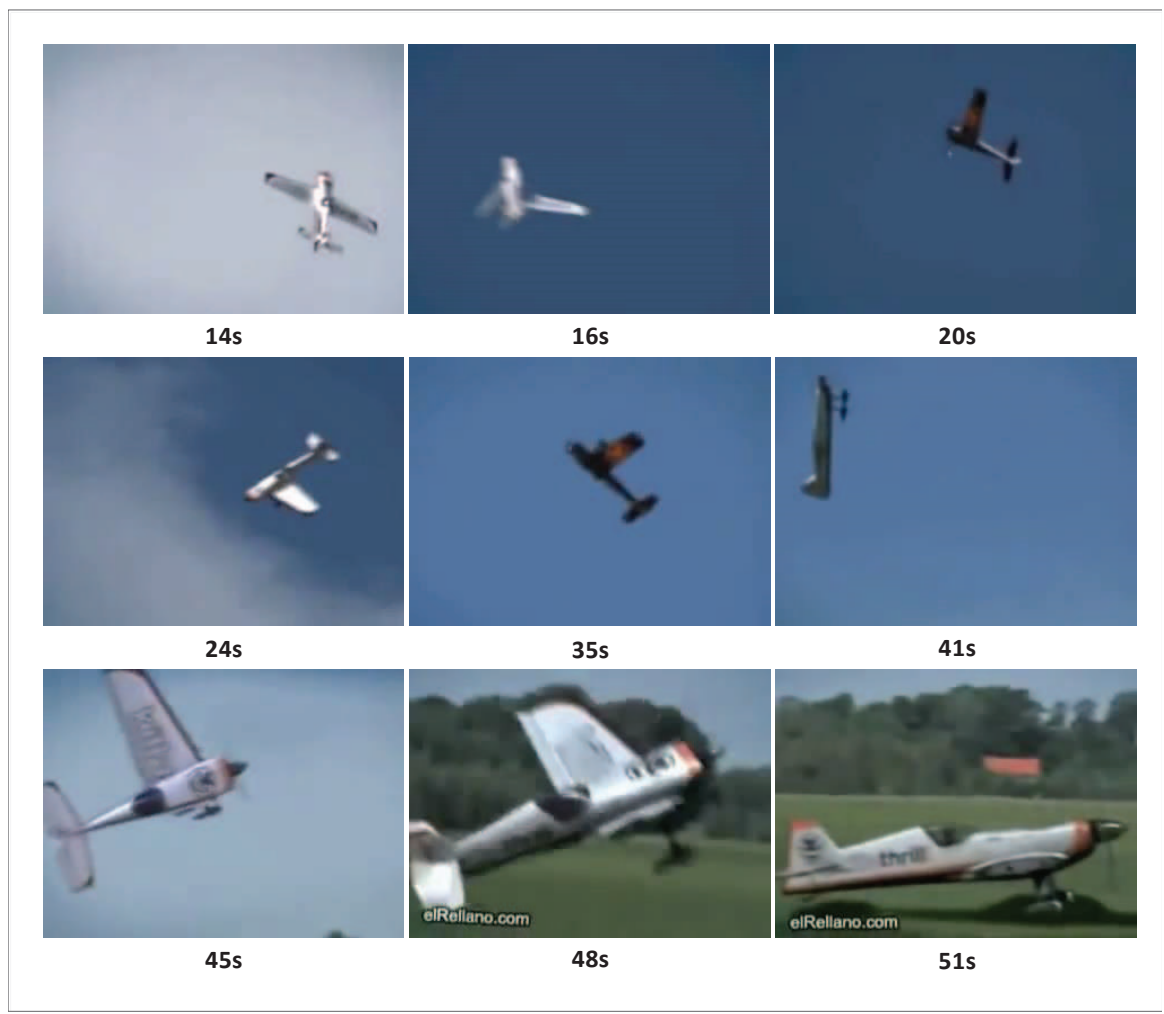} \caption{A
fixed-wing aircraft lost its right wing in the air}%
\label{fixdwingfailure}%
\end{figure}

The pilot, we guess, did not have such an experience in dealing with the
accident, but the pilot landed the damaged aircraft safely. This is similar to
another accident on May 1, 1983, where an Israeli Air Force (IAF) F-15 Baz
(actually the F-15D No. 957) collided with an A-4 Skyhawk in an air combat
training mission over the Negev\footnote{https://theaviationgeekclub.com}.
However, the two accidents are very different. The Israeli pilot was able to
maintain control because of the lift generated by the large areas of the
fuselage, stabilators, and remaining wing. However, the pilot in Fig.1 could
not control the aircraft's attitude anymore.\begin{figure}[h]
\centering \includegraphics[scale= 0.5]{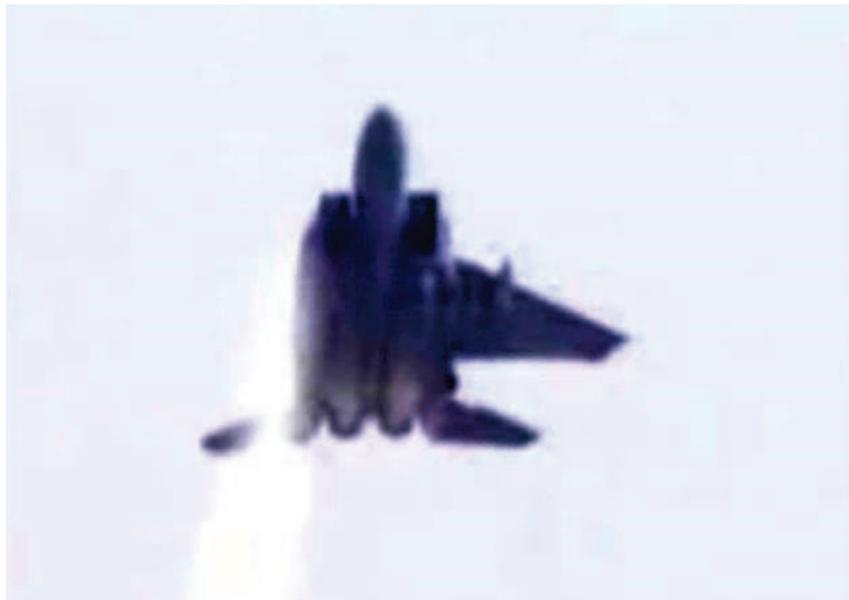}\caption{A F-15 lost
its most right-wing in the air}%
\end{figure}

From the two accidents, we can learn three control properties from
sophisticated pilots:

\begin{itemize}
\item \textbf{Property 1}. Such a failure is not defined and analyzed a
priori. In practice, it is impossible and tedious to determine all failures.

\item \textbf{Property 2}. FDD is unavailable. In the accident shown in Fig.2,
the pilot was later quoted as saying, \textquotedblleft(I) probably would have
ejected if I knew what had happened\textquotedblright. In practice, some
faults are also unobservable. On the other hand, this implies that the
impaired model is unavailable for controllers. An approach for dealing with
this is to automatically optimize or reshape the trajectory of an impaired
aircraft for particular tasks in a way that takes any impairment into account
\cite{Steinberg(2015)}. However, this method heavily depends on the impaired model.

\item \textbf{Property 3}. More importantly, although the damage caused the
aircraft's attitude to be uncontrollable in the accident shown in Fig.1, the
damaged aircraft was landed as safely as possible. As the matter current is
either passive or active FTC, the faulty system is often assumed to state
controllable as the same as the original system implicitly. To clarify, if a
fault causes the original system's state to be uncontrollable, then we call it
a `\emph{disaster}'. Under disasters, pilots always try their best to land a
damaged aircraft within a minimal time, rather than care whether the aircraft
is controllable.
\end{itemize}

The three properties motivate us to find the control principle behind it. With
the principle, we aim to develop a new flight-control framework with
\textbf{Properties 1-3} to improve the reliability and safety of an aircraft.

In this paper, the \emph{gravity-compensation-first principle} is proposed
from the \emph{first principle}%
\footnote{https://en.wikipedia.org/wiki/First\_principle}. We will formulate
the principle in three aspects: force, impulse, and energy. With the
principle, we study passive disaster-tolerant control of multicopters and
fixed-wing using the model predictive control approach.

\section{Gravity-Compensation-First Principle}

\subsection{Gravity-Compensation-First Principle}

According to Newton's second law, we have%
\begin{align}
\mathbf{\dot{p}}  &  =\mathbf{v}\nonumber\\
\mathbf{\dot{v}}  &  =\mathbf{g+}\frac{1}{m}\mathbf{f} \label{secondlaw}%
\end{align}
where $\mathbf{p,v}\in\mathbf{%
\mathbb{R}
}^{3}$ are position and velocity of an aircraft, $\mathbf{g}\in\mathbf{%
\mathbb{R}
}^{3}$ is acceleration of gravity, $m\in\mathbf{%
\mathbb{R}
}$ is mass of the aircraft, $\mathbf{f}\in\mathbf{\mathcal{F}}\subset\mathbf{%
\mathbb{R}
}^{3}$ is the force on the aircraft other than gravity. As shown in Fig.3, the
function of the nose of the aircraft pointing upward aims at \emph{decreasing
falling speed}, which is a significant change the pilot made around 35s. After
that, the falling speed is under control. This is a crucial step to why the
pilot can survive the accident. Also, this implies that \emph{compensating for
gravity} is the first requirement. \begin{figure}[h]
\centering \includegraphics[scale= 0.6]{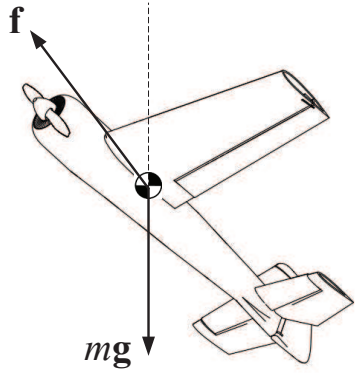}\caption{The nose
of the aircraft pointing upward to compensate for gravity}%
\end{figure}

The desired acceleration is set $\mathbf{a}_{d}\in\mathbf{%
\mathbb{R}
}^{3}$. With it, (\ref{secondlaw}) becomes%
\begin{align}
\mathbf{\dot{p}}  &  =\mathbf{v}\nonumber\\
\mathbf{\dot{v}}  &  =\mathbf{a}_{d}. \label{disiredsecondlaw}%
\end{align}
If the aircraft is required to track $\mathbf{p}_{d}\in\mathbf{%
\mathbb{R}
}^{3},$ then a straightforward design of the desired acceleration is%
\begin{equation}
\mathbf{a}_{d}=-k_{p}\left(  \mathbf{p}-\mathbf{p}_{d}\right)  -k_{d}\left(
\mathbf{v}-\mathbf{\dot{p}}_{d}\right)  \label{desiredaccel}%
\end{equation}
where $k_{p},k_{d}>0.$ So, compared (\ref{disiredsecondlaw}) with
(\ref{secondlaw}), we expect that the \emph{desired} force of $\mathbf{f}$
denoted by $\mathbf{f}_{d}$ is%
\begin{equation}
\mathbf{a}_{d}=\mathbf{g+}\frac{1}{m}\mathbf{f}_{d}. \label{desiredacc}%
\end{equation}
Then%
\begin{equation}
\mathbf{f}_{d}=\underset{\text{Compensating for gravity}%
}{\underbrace{-m\mathbf{g}}}+\underset{\text{Tracking desired trajectory}%
}{\underbrace{m\mathbf{a}_{d}}} \label{forcedecomposition}%
\end{equation}
From (\ref{forcedecomposition}), It is apparent that the control is decomposed
into two tasks: \emph{compensating for gravity} and \emph{tracking the desired
trajectory}. Traditionally, the two tasks are mixed together by simple
addition. Sometimes, it is impossible to satisfy the two objectives
simultaneously because the force $\mathbf{f}$ is confined. A trade-off has to
be made. From the accident shown in Fig.1, the task of compensating for
gravity should be dominated. So, the gravity-compensation-first principle is obtained.

\textbf{Gravity-Compensation-First Principle}: \emph{All control gives
priority to compensating for the effect caused by the gravity of the aircraft,
with the left authority used to meet the requirements of the desired motion of
the aircraft.}

For a single aircraft, the only danger is hitting the ground. If the gravity
of the aircraft is compensated for totally, then the aircraft becomes a
spacecraft in outer space without worrying about falling to the ground.
According to the \emph{first principle}, we only stop at the resultant force
based on Newton's second law to study the problem rather than going deep into
torques and further actuators. After compensating for gravity, the left
control will be used to realize the trajectory tracking. In the following, we
will formulate the principle in three aspects: force, impulse, and energy.

\textbf{(1) Force Viewpoint}

Let $\mathbf{f}_{g},\mathbf{f}_{t}$ be used to compensate for gravity and
track the desired trajectory, respectively. According to the
\textit{gravity-compensation-first principle}, the procedure to obtain a new
$\mathbf{f}_{d}$ is divided into three steps.

\begin{itemize}
\item Step 1. Obtain the feasible compensate-for-gravity term with \emph{the
first priority} as%
\begin{equation}
\mathbf{f}_{g}^{\ast}=\underset{\mathbf{f}_{g}\in\mathcal{F}\subset\mathcal{%
\mathbb{R}
}^{3}}{\arg\min}\left\Vert \mathbf{f}_{g}-m\mathbf{g}\right\Vert .
\label{compensate-for-gravity_force}%
\end{equation}

\end{itemize}

Step 2. Obtain the feasible tracking-desired-trajectory term
\emph{secondarily} as%
\begin{equation}
\mathbf{f}_{t}^{\ast}=\underset{-\mathbf{f}_{g}^{\ast}+\mathbf{f}_{t}%
\in\mathcal{F}\subset\mathcal{%
\mathbb{R}
}^{3}}{\arg\min}\left\Vert \mathbf{f}_{g}-m\mathbf{a}_{d}\right\Vert .
\label{tracking-desired-trajectory_force}%
\end{equation}

\begin{itemize}
\item Step 3. Obtain the synthesis term as%
\begin{equation}
\mathbf{f}_{d}=-\mathbf{f}_{g}^{\ast}+\mathbf{f}_{t}^{\ast}. \label{synthesis}%
\end{equation}

\end{itemize}

\textbf{Example 1}. Suppose $\mathcal{F}=\left\{  \left.  \mathbf{x}%
\right\vert \left\Vert \mathbf{x}\right\Vert \leq15\right\}  $ and
\[
m\mathbf{g}=\left[
\begin{array}
[c]{c}%
0\\
0\\
9.8
\end{array}
\right]  ,m\mathbf{a}_{d}=\left[
\begin{array}
[c]{c}%
20\\
20\\
0
\end{array}
\right]  .
\]
Then, according to \textit{Step 1}, we have%
\[
\mathbf{f}_{g}^{\ast}=\left[
\begin{array}
[c]{c}%
0\\
0\\
9.8
\end{array}
\right]  .
\]
Furthermore, according to \textit{Step 2}, we have%
\[
\mathbf{f}_{t}^{\ast}=\left[
\begin{array}
[c]{c}%
11.4\left/  \sqrt{2}\right. \\
11.4\left/  \sqrt{2}\right. \\
0
\end{array}
\right]  .
\]
Therefore
\[
\mathbf{f}_{d}=-\underset{\mathbf{f}_{g}^{\ast}}{\underbrace{\left[
\begin{array}
[c]{c}%
0\\
0\\
9.8
\end{array}
\right]  }}+\underset{\mathbf{f}_{t}^{\ast}}{\underbrace{\left[
\begin{array}
[c]{c}%
11.4\left/  \sqrt{2}\right. \\
11.4\left/  \sqrt{2}\right. \\
0
\end{array}
\right]  }}.
\]

According to the \textit{gravity-compensation-first principle}, another
approximate but concise procedure to obtain $\mathbf{f}_{d}$ is given as%
\begin{equation}%
\begin{array}
[c]{ll}%
\underset{\mathbf{f}_{g},\mathbf{f}_{t}\in\mathcal{%
\mathbb{R}
}^{3}}{\min} & w_{g}\left\Vert \mathbf{f}_{g}-m\mathbf{g}\right\Vert
^{2}+w_{t}\left\Vert \mathbf{f}_{t}-m\mathbf{a}_{d}\right\Vert ^{2}\\
\text{s.t.} & -\mathbf{f}_{g}+\mathbf{f}_{t}\in\mathcal{F}\subset\mathcal{%
\mathbb{R}
}^{3}%
\end{array}
. \label{Force}%
\end{equation}
where $w_{g}\gg w_{t}>0$ implies \emph{gravity-compensation first}. After
obtaining the solutions $\mathbf{f}_{g}^{\ast},\mathbf{f}_{t}^{\ast}$\ to the
optimization (\ref{Force}), we can synthesize them as (\ref{synthesis}). By
using the new procedure, it is easy to obtain control by one optimal problem.
We will recommend it to formulate the principle from the \emph{impulse
viewpoint} and the \emph{energy viewpoint}.

\textbf{(2) Impulse Viewpoint}

From the impulse viewpoint, we have%
\[
\mathbf{f}_{g}^{\ast}=\underset{\mathbf{f}_{g}\in\mathcal{F}\subset
\mathcal{C}\left(  \mathcal{%
\mathbb{R}
}^{3},\left[  t_{0},t_{0}+T\right]  \right)  }{\arg\min}\left\Vert \int%
_{t_{0}}^{t_{0}+T}\left(  \mathbf{f}_{g}\left(  t\right)  -m\mathbf{g}\right)
\text{d}t\right\Vert .
\]
After that, we further have%
\[
\mathbf{f}_{t}^{\ast}=\underset{\mathbf{f}_{t}-\mathbf{f}_{g}^{\ast}%
\in\mathcal{F}\subset\mathcal{C}\left(  \mathcal{%
\mathbb{R}
}^{3},\left[  t_{0},t_{0}+T\right]  \right)  }{\arg\min}\left\Vert \int%
_{t_{0}}^{t_{0}+T}\left(  \mathbf{f}_{t}\left(  t\right)  -m\mathbf{a}%
_{d}\left(  t\right)  \right)  \text{d}t\right\Vert .
\]

Unlike the optimization in the force viewpoint, the optimization here aims at
finding a continuous vector $\mathbf{f}_{g}^{\ast}\left(  t\right)
\in\mathcal{%
\mathbb{R}
}^{3},$ $t\in\left[  t_{0},t_{0}+T\right]  $. Obviously, it has infinite
number of solutions. So, it is a bit early to determine $\mathbf{f}_{g}^{\ast
}$ first then $\mathbf{f}_{t}^{\ast}$. The variables $\mathbf{f}_{g}^{\ast}$
and $\mathbf{f}_{t}^{\ast}$ are suggested to combine together to complete the
optimization. What is more, some constraints could be put on, such as the
minimum energy or minimum change. We has%
\begin{equation}%
\begin{array}
[c]{ll}%
\underset{\mathbf{f}_{g},\mathbf{f}_{t}\in\mathcal{C}\left(  \mathcal{%
\mathbb{R}
}^{3},\left[  t_{0},t_{0}+T\right]  \right)  }{\min} &
\begin{array}
[c]{l}%
w_{g}\left\Vert \int_{t_{0}}^{t_{0}+T}\left(  \mathbf{f}_{g}\left(  t\right)
-m\mathbf{g}\right)  \text{d}t\right\Vert ^{2}+w_{t}\left\Vert \int_{t_{0}%
}^{t_{0}+T}\left(  \mathbf{f}_{t}\left(  t\right)  -m\mathbf{a}_{d}\left(
t\right)  \right)  \text{d}t\right\Vert ^{2}\\
+w_{e}\int_{t_{0}}^{t_{0}+T}\left\Vert -\mathbf{f}_{g}\left(  t\right)
+\mathbf{f}_{t}\left(  t\right)  \right\Vert ^{2}\text{d}t
\end{array}
\\
\text{s.t.} & -\mathbf{f}_{g}+\mathbf{f}_{t}\in\mathcal{F}\subset
\mathcal{C}\left(  \mathcal{%
\mathbb{R}
}^{3},\left[  t_{0},t_{0}+T\right]  \right)
\end{array}
\label{Impulse}%
\end{equation}
where $w_{g}\gg w_{t}\gg w_{e}>0.$ The additional term $w_{e}\int_{t_{0}%
}^{t_{0}+T}\left\Vert -\mathbf{f}_{g}\left(  t\right)  +\mathbf{f}_{t}\left(
t\right)  \right\Vert ^{2}$d$t$ is used to find a solution satisfying the
gravity compensation first and tracking task secondarily with a minimum
energy. In this viewpoint, we do not need $\mathbf{f}_{g}\left(  t\right)
=m\mathbf{g},$ $\mathbf{f}_{t}\left(  t\right)  =m\mathbf{a}_{d}\left(
t\right)  $ at each time during $t\in\left[  t_{0},t_{0}+T\right]  $. After
obtaining the solutions $\mathbf{f}_{g}^{\ast},\mathbf{f}_{t}^{\ast}$\ to the
optimization (\ref{Impulse}), we can synthesize them as (\ref{synthesis}).
This optimization can be also put into the framework of the model predictive
control \cite{Rakovic(2018)}, where the cost function over a receding horizon
$t\in\left[  t_{0},t_{0}+T\right]  $ is defined like (\ref{Impulse}).

(3) \textbf{Energy Viewpoint}

When a helicopter's main rotor fails, it will spin down. To save the
helicopter, experienced pilots first increase the rotor rotation speed and
finally increase the collective pitch at a distance from the ground to convert
the rotor kinetic energy into work in the opposite direction of the
helicopter's gravity to minimize the falling kinetic energy effect caused by
gravity. This process includes an energy storage process, which does not
compensate for gravity all the time. Assume an aircraft lands from an altitude
$\mathbf{h}_{\text{0}}=\left[
\begin{array}
[c]{ccc}%
0 & 0 & h_{0}%
\end{array}
\right]  ^{\text{T}}$ to $\mathbf{h}_{\text{1}}=\left[
\begin{array}
[c]{ccc}%
0 & 0 & h_{0}+H
\end{array}
\right]  ^{\text{T}},$ $H>0$ (z-axis points perpendicularly to the ground).
Inspired by this, the optimization objective can reflect the
gravity-compensation-first principle in terms of energy as%
\begin{equation}%
\begin{array}
[c]{ll}%
\underset{\mathbf{f}_{g},\mathbf{f}_{t}\in\mathcal{C}\left(  \mathcal{%
\mathbb{R}
}^{3},\left[  h_{0},h_{0}+H\right]  \right)  }{\min} &
\begin{array}
[c]{l}%
w_{g}\left\Vert \int_{\mathbf{h}_{\text{0}}}^{\mathbf{h}_{\text{1}}}\left(
\mathbf{f}_{g}\left(  h\right)  -m\mathbf{g}\right)  ^{\text{T}}%
\text{d}\mathbf{h}\right\Vert ^{2}+w_{t}\left\Vert \int_{\mathbf{h}_{\text{0}%
}}^{\mathbf{h}_{\text{1}}}\left(  \mathbf{f}_{t}\left(  h\right)
-m\mathbf{a}_{d}\left(  h\right)  \right)  ^{\text{T}}\text{d}\mathbf{h}%
\right\Vert ^{2}\\
+w_{e}\left\Vert \int_{\mathbf{h}_{0}}^{\mathbf{h}_{1}}\left(  -\mathbf{f}%
_{g}\left(  h\right)  +\mathbf{f}_{t}\left(  h\right)  \right)  ^{\text{T}%
}\text{d}\mathbf{h}\right\Vert ^{2}%
\end{array}
\\
\text{s.t.} & -\mathbf{f}_{g}+\mathbf{f}_{t}\in\mathcal{F}\subset
\mathcal{C}\left(  \mathcal{%
\mathbb{R}
}^{3},\left[  h_{0},h_{0}+H\right]  \right)  ,\mathbf{h=}\left[
\begin{array}
[c]{ccc}%
0 & 0 & h
\end{array}
\right]  ^{\text{T}}%
\end{array}
\label{Energy}%
\end{equation}
where $w_{g}\gg w_{t}\gg w_{e}>0$, the additional term $\left\Vert
\int_{\mathbf{h}_{0}}^{\mathbf{h}_{1}}\left(  -\mathbf{f}_{g}+\mathbf{f}%
_{t}\right)  ^{\text{T}}\text{d}\mathbf{h}\right\Vert ^{2}$\ is used to find a
solution satisfying the gravity compensation first and tracking task
secondarily with a minimum energy. In this viewpoint, we also do not need
$\mathbf{f}_{g}\left(  h\right)  =m\mathbf{g},$ $\mathbf{f}_{t}\left(
h\right)  =m\mathbf{a}_{d}\left(  h\right)  $ at each height during
$h\in\left[  h_{0},h_{0}+H\right]  $. This optimization can be also put into
the framework of the model predictive control \cite{Rakovic(2018)}, where the
cost function over a receding horizon $h\in\left[  h_{0},h_{0}+H\right]  $ is
defined like (\ref{Energy}).

\subsection{Gravity-Compensation-First Principle in the Presence of
Disturbances}

In the presence of disturbances, (\ref{secondlaw}) is rewritten as%
\begin{align}
\mathbf{\dot{p}}  &  =\mathbf{v}\nonumber\\
\mathbf{\dot{v}}  &  =\mathbf{g+}\frac{1}{m}\mathbf{f+}\frac{1}{m}\mathbf{d}
\label{secondlaw1}%
\end{align}
where $\mathbf{d=[}d_{1}$ $d_{2}$ $d_{3}]^{\text{T}}\in\mathbf{%
\mathbb{R}
}^{3}$ is a disturbance supposed to be estimated exactly. In order to achieve
the objective (\ref{disiredsecondlaw}), we should have%
\begin{equation}
\mathbf{f}_{d}=\underset{\text{Compensating for gravity}%
}{\underbrace{-m\mathbf{g}}}+\underset{\text{Compensating for disturbance}%
}{\underbrace{-\mathbf{d}}}+\underset{\text{Tracking desired trajectory}%
}{\underbrace{m\mathbf{a}_{d}}}. \label{forcedecomposition1}%
\end{equation}

From (\ref{forcedecomposition}), it is very clear that the control is
decomposed into three tasks, namely \textit{compensating for gravity,
compensating for disturbance} and \textit{tracking desired trajectory}.
Traditionally, the three tasks are mixed together by simple addition. Or,
according to the \textit{gravity-compensation-first principle},\textbf{ }the
second and third tasks are mixed together. If so, it is inappropriate because
the disturbance may also make the aircraft fall. For example, the disturbance
is a payload attached to the aircraft. Following a similar idea, we decompose
the disturbance to be the sum of two perpendicular vector components as%
\[
\mathbf{d=}\underset{\mathbf{d}_{g}}{\underbrace{\frac{\mathbf{g}^{\text{T}%
}\mathbf{d}}{\mathbf{g}^{\text{T}}\mathbf{g}}\mathbf{g}}}+\underset{\mathbf{d}%
-\mathbf{d}_{g}}{\underbrace{\left(  \mathbf{d-}\frac{\mathbf{g}^{\text{T}%
}\mathbf{d}}{\mathbf{g}^{\text{T}}\mathbf{g}}\mathbf{g}\right)  }}%
\]
where $\mathbf{d}_{g}$ is a component along the direction of $\mathbf{g.}$ If
$\mathbf{g=[}0$ $0$ $g]^{\text{T}},$ then $\mathbf{d}_{g}=\mathbf{[}0$ $0$
$d_{3}]^{\text{T}}.$ Furthermore, it is not easy to order the priorities of
the tasks of the \textit{compensating-for-disturbance other than gravity
direction }and\textit{ tracking desired trajectory. }Therefore,\textit{ }we
write $\mathbf{d}_{g}$ into the compensating-for-gravity term, and put
$\mathbf{d}-\mathbf{d}_{g}$ to the tracking-desired-trajectory term as%
\begin{equation}
\mathbf{f}_{d}=\underset{\text{Compensating for gravity}%
}{\underbrace{-m\mathbf{g-d}_{g}}}+\underset{\text{Tracking desired
trajectory}}{\underbrace{m\mathbf{a}_{d}-\left(  \mathbf{d}-\mathbf{d}%
_{g}\right)  }}. \label{forcedecomposition2}%
\end{equation}

Similarly, we can also obtain three optimization problems in three aspects:
force, impulse, and energy.

\begin{itemize}
\item \textbf{Force. }The optimization problem (\ref{Force}) is rewritten as%
\begin{equation}%
\begin{array}
[c]{ll}%
\underset{\mathbf{f}_{g},\mathbf{f}_{t}\in\mathcal{%
\mathbb{R}
}^{3}}{\min} & w_{g}\left\Vert \mathbf{f}_{g}\left(  t\right)  -m\mathbf{g-d}%
_{g}\left(  t\right)  \right\Vert ^{2}+w_{t}\left\Vert \mathbf{f}_{t}\left(
t\right)  -m\mathbf{a}_{d}\left(  t\right)  -\left(  \mathbf{d}\left(
t\right)  -\mathbf{d}_{g}\left(  t\right)  \right)  \right\Vert ^{2}\\
\text{s.t.} & -\mathbf{f}_{g}+\mathbf{f}_{t}\in\mathcal{F}\subset\mathcal{%
\mathbb{R}
}^{3}%
\end{array}
. \label{force1}%
\end{equation}

\item \textbf{Impulse. }The optimization problem (\ref{Impulse}) is rewritten
as%
\begin{equation}%
\begin{array}
[c]{ll}%
\underset{\mathbf{f}_{g},\mathbf{f}_{t}\in\mathcal{C}\left(  \mathcal{%
\mathbb{R}
}^{3},\left[  t_{0},t_{0}+T\right]  \right)  }{\min} &
\begin{array}
[c]{l}%
w_{g}\left\Vert \int_{t_{0}}^{t_{0}+T}\left(  \mathbf{f}_{g}\left(  t\right)
-m\mathbf{g-d}_{g}\left(  t\right)  \right)  \text{d}t\right\Vert ^{2}\\
+w_{t}\left\Vert \int_{t_{0}}^{t_{0}+T}\left(  \mathbf{f}_{t}\left(  t\right)
-m\mathbf{a}_{d}\left(  t\right)  -\left(  \mathbf{d}\left(  t\right)
-\mathbf{d}_{g}\left(  t\right)  \right)  \right)  \text{d}t\right\Vert ^{2}\\
+w_{e}\int_{t_{0}}^{t_{0}+T}\left\Vert -\mathbf{f}_{g}\left(  t\right)
+\mathbf{f}_{t}\left(  t\right)  \right\Vert ^{2}\text{d}t
\end{array}
\\
\text{s.t.} & -\mathbf{f}_{g}+\mathbf{f}_{t}\in\mathcal{F}\subset
\mathcal{C}\left(  \mathcal{%
\mathbb{R}
}^{3},\left[  t_{0},t_{0}+T\right]  \right)
\end{array}
. \label{Impulse1}%
\end{equation}

\item \textbf{Energy. }The optimization problem (\ref{Energy}) is rewritten as%
\begin{equation}%
\begin{array}
[c]{ll}%
\underset{\mathbf{f}_{g},\mathbf{f}_{t}\in\mathcal{C}\left(  \mathcal{%
\mathbb{R}
}^{3},\left[  h_{0},h_{0}+H\right]  \right)  }{\min} &
\begin{array}
[c]{l}%
w_{g}\left\Vert \int_{\mathbf{h}_{\text{0}}}^{\mathbf{h}_{\text{1}}}\left(
\mathbf{f}_{g}\left(  h\right)  -m\mathbf{g-d}_{g}\left(  h\right)  \right)
^{\text{T}}\text{d}\mathbf{h}\right\Vert ^{2}\\
+w_{t}\left\Vert \int_{\mathbf{h}_{\text{0}}}^{\mathbf{h}_{\text{1}}}\left(
\mathbf{f}_{t}\left(  h\right)  -m\mathbf{a}_{d}\left(  h\right)  -\left(
\mathbf{d}\left(  h\right)  -\mathbf{d}_{g}\left(  h\right)  \right)  \right)
^{\text{T}}\text{d}\mathbf{h}\right\Vert ^{2}\\
+w_{e}\left\Vert \int_{\mathbf{h}_{0}}^{\mathbf{h}_{1}}\left(  -\mathbf{f}%
_{g}\left(  h\right)  +\mathbf{f}_{t}\left(  h\right)  \right)  ^{\text{T}%
}\text{d}\mathbf{h}\right\Vert ^{2}%
\end{array}
\\
\text{s.t.} & -\mathbf{f}_{g}+\mathbf{f}_{t}\in\mathcal{F}\subset
\mathcal{C}\left(  \mathcal{%
\mathbb{R}
}^{3},\left[  h_{0},h_{0}+H\right]  \right)  ,\mathbf{h=}\left[
\begin{array}
[c]{ccc}%
0 & 0 & h
\end{array}
\right]  ^{\text{T}}%
\end{array}
. \label{Energy1}%
\end{equation}

\end{itemize}

After obtaining the optimal solutions $\mathbf{f}_{g}^{\ast},\mathbf{f}%
_{t}^{\ast}$\ to these optimizations above, we can synthesize them as
(\ref{synthesis}).

\section{Applications}

\subsection{Coordinate System}

The Earth-fixed coordinate frame $o_{e}x_{e}y_{e}z_{e}$ is used to study an
aircraft's dynamic states relative to the Earth's surface and to determine its
three-dimensional (3D) position. The Earth's curvature is ignored, namely the
Earth's surface is assumed to be flat. The initial position of the aircraft or
the center of the Earth is often set as the coordinate origin $o_{e}$, the
$o_{e}x_{e}$ axis points to a certain direction in the horizontal plane, and
the $o_{e}z_{e}$ axis points perpendicularly to the ground. Then, the
$o_{e}y_{e}$ axis is determined according to the right-hand rule.

The aircraft-body coordinate frame $o_{b}x_{b}y_{b}z_{b}$ is fixed to an
aircraft. The center of gravity of the aircraft is chosen as the origin
$o_{b}$ of $o_{b}x_{b}y_{b}z_{b}$. The $o_{b}x_{b}$ axis points to the nose
direction in the symmetric plane of the aircraft (nose direction is related to
the plus-configuration multicopter or the X-configuration multicopter). The
$o_{b}z_{b}$ axis is in the symmetric plane of the aircraft, pointing
downward, perpendicular to the $o_{b}x_{b}$ axis. The $o_{b}y_{b}$ axis is
determined according to the right-hand rule. The rotation matrix $\mathbf{R}$
can map the vector in the aircraft-body coordinate frame to the Earth-fixed
coordinate frame.

The origin $o_{w}$ of the wind coordinate frame $o_{w}x_{w}y_{w}z_{w}$ is also
at the center of gravity of an aircraft. The $o_{w}x_{w}$ axis is aligned with
the airspeed vector, and points to the front. The $o_{w}z_{w}$ axis is in the
symmetric plane of the aircraft, pointing downward, perpendicular to the
$o_{w}x_{w}$ axis, and the $o_{w}y_{w}$ axis is determined according to the
right-hand rule. The angle of attack $\alpha$ is defined as the angle between
the $o_{b}x_{b}$ axis and the projection vector of the $o_{w}x_{w}$ axis onto
the plane $o_{b}x_{b}z_{b},$ while the sideslip angle $\beta$ is the angle
between the $o_{w}x_{w}$ axis and the plane $o_{b}x_{b}z_{b}.$ The rotation
matrix $\mathbf{R}_{a}\left(  \alpha,\beta\right)  $ can map the vector in the
wind coordinate frame to the aircraft-body coordinate frame.

\subsection{Application to a Quadcopter Control}

\begin{figure}[h]
\centering \includegraphics[scale= 0.6]{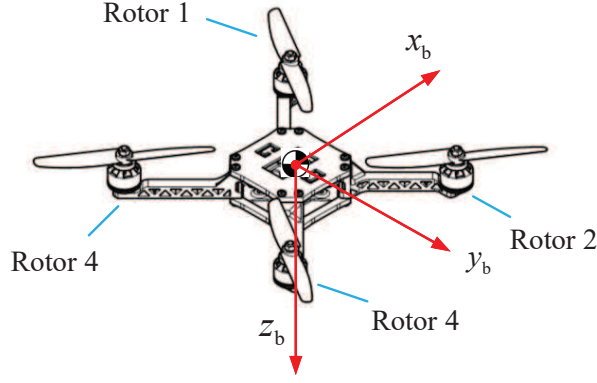}\caption{A quadcopter}%
\end{figure}

As shown in Fig.4, a quadcopter's dynamic model is expressed
as\cite{Quan(2017)}%
\begin{align}
\mathbf{\dot{p}}  &  =\mathbf{v}\nonumber\\
\mathbf{\dot{v}}  &  =\mathbf{g-}\frac{1}{m}f\mathbf{R}\left(  \mathbf{q}%
\right)  \mathbf{e}_{3}\nonumber\\
\mathbf{\dot{q}}  &  =\left[
\begin{array}
[c]{cc}%
0 & -\boldsymbol{\omega}^{\text{T}}\\
\boldsymbol{\omega} & -\left[  \boldsymbol{\omega}\right]  _{\times}%
\end{array}
\right]  \mathbf{q}\nonumber\\
\mathbf{J}\boldsymbol{\dot{\omega}}  &  =-\boldsymbol{\omega}\times\left(
\mathbf{J}\boldsymbol{\omega}\right)  +\mathbf{G}+\boldsymbol{\tau}
\label{quadcopter}%
\end{align}
with%
\begin{equation}
\left[
\begin{array}
[c]{c}%
f\\
\tau_{x}\\
\tau_{y}\\
\tau_{z}%
\end{array}
\right]  =\underset{\mathbf{H}}{\underbrace{\left[
\begin{array}
[c]{cccc}%
1 & 1 & 1 & 1\\
\frac{\sqrt{2}}{2}d & -\frac{\sqrt{2}}{2}d & \frac{\sqrt{2}}{2}d &
-\frac{\sqrt{2}}{2}d\\
\frac{\sqrt{2}}{2}d & \frac{\sqrt{2}}{2}d & -\frac{\sqrt{2}}{2}d &
-\frac{\sqrt{2}}{2}d\\
c_{M}\left/  c_{T}\right.  & -c_{M}\left/  c_{T}\right.  & c_{M}\left/
c_{T}\right.  & -c_{M}\left/  c_{T}\right.
\end{array}
\right]  }}\underset{\mathbf{T}}{\underbrace{\left[
\begin{array}
[c]{c}%
T_{1}\\
T_{2}\\
T_{3}\\
T_{4}%
\end{array}
\right]  }} \label{quadeffic}%
\end{equation}
where $\mathbf{p}\in\mathbf{%
\mathbb{R}
}^{3}$ and $\mathbf{v}\in\mathbf{%
\mathbb{R}
}^{3}$ are position and velocity of the center of the multicopter in frame
$o_{e}x_{e}y_{e}z_{e}$, respectively; the gravity vector is $\mathbf{g=[}0$
$0$ $g\mathbf{]}^{\text{T}}\in\mathbf{%
\mathbb{R}
}^{3}$, the mass of the multicopter is $m>0;$ $\mathbf{R}\left(
\mathbf{q}\right)  \ $is a rotation matrix, determined by a quaternion
$\mathbf{q}\in\mathbf{%
\mathbb{R}
}^{4}$; $\boldsymbol{\omega}\in\mathbf{%
\mathbb{R}
}^{3}$ is the angular velocity of the multicopter expressed in frame
$o_{b}x_{b}y_{b}z_{b};$ $\mathbf{J}\in\mathbf{%
\mathbb{R}
}^{3\times3}\ $represents the multicopter moment of inertia; $\mathbf{G}$
represents the gyroscopic torque of rotors and propellers; $f$ $>0$ represents
the magnitude of the total propeller thrust, $\boldsymbol{\tau=[}\tau_{x}$
$\tau_{y}$ $\tau_{z}\boldsymbol{]}^{\text{T}}\in\mathbf{%
\mathbb{R}
}^{3}$ represents the moments generated by the propellers in the aircraft-body
coordinate frame $o_{b}x_{b}y_{b}z_{b};$ $d$ $>0$ represents the distance
between the body center and any motor; $c_{M},c_{T}>0$ are coefficients
related to the pair of a propeller and motor; $T_{i}\in\lbrack0,T_{m}]$ are
the propeller thrust with a maximum $T_{m}>0,$ $i=1,2,3,4.$ The objective is
to design $T_{i}$ to make the multicopter track $\mathbf{p}_{d}\left(
t\right)  $\ as accurately as possible in the case of both no failure and the
first motor failing completely without loss of generality.

By using (\ref{desiredacc}), we hope the desired total propeller thrust and
desired rotation matrix to satisfy%
\[
\mathbf{g-}\frac{1}{m}f_{d}\mathbf{R}\left(  \mathbf{q}_{d}\right)
\mathbf{e}_{3}=\mathbf{a}_{d}%
\]
where $f_{d},\mathbf{q}_{d}$ are finally determined by the propeller thrust
$T_{i}\in\lbrack0,T_{m}].$ Since
\begin{equation}
\mathbf{f}_{d}=-f_{d}\mathbf{R}\left(  \mathbf{q}_{d}\right)  \mathbf{e}_{3}%
\end{equation}
we can obtain that%
\begin{equation}%
\begin{array}
[c]{ll}%
\underset{T_{id}\in\mathcal{C}\left(  \left[  0,T_{m}\right]  ,\left[
t_{0},t_{0}+T\right]  \right)  ,i=1,2,3,4}{\min} &
\begin{array}
[c]{l}%
w_{g}\left\Vert \int_{t_{0}}^{t_{0}+T}\left(  \mathbf{f}_{g}\left(  t\right)
-m\mathbf{g}\right)  \text{d}t\right\Vert ^{2}+w_{t}\left\Vert \int_{t_{0}%
}^{t_{0}+T}\left(  \mathbf{f}_{t}\left(  t\right)  -m\mathbf{a}_{d}\left(
t\right)  \right)  \text{d}t\right\Vert ^{2}\\
+w_{e}\int_{t_{0}}^{t_{0}+T}\left\Vert -\mathbf{f}_{g}\left(  t\right)
+\mathbf{f}_{t}\left(  t\right)  \right\Vert ^{2}\text{d}t
\end{array}
\\
\text{s.t.} &
\begin{array}
[c]{lll}%
f_{d}\mathbf{R}\left(  \mathbf{q}_{d}\right)  \mathbf{e}_{3} & = &
-\mathbf{f}_{g}+\mathbf{f}_{t}\\
\mathbf{\dot{q}}_{d} & = & \left[
\begin{array}
[c]{cc}%
0 & -\boldsymbol{\omega}_{d}^{\text{T}}\\
\boldsymbol{\omega}_{d} & -\left[  \boldsymbol{\omega}_{d}\right]  _{\times}%
\end{array}
\right]  \mathbf{q}_{d}\\
\mathbf{J}\boldsymbol{\dot{\omega}}_{d} & = & -\boldsymbol{\omega}_{d}%
\times\left(  \mathbf{J}\boldsymbol{\omega}_{d}\right)  +\mathbf{G}%
+\boldsymbol{\tau}_{d}\\
\left[
\begin{array}
[c]{c}%
f_{d}\\
\boldsymbol{\tau}_{d}%
\end{array}
\right]  & = & \mathbf{HT}_{d},\mathbf{T}_{d}=[T_{1d}\text{ }T_{2d}\text{
}T_{3d}\text{ }T_{4d}]^{\text{T}}%
\end{array}
.
\end{array}
\end{equation}
according to the gravity-compensation-first principle from the impulse
viewpoint. Furthermore, if the first motor fails completely, then%
\begin{equation}%
\begin{array}
[c]{ll}%
\underset{T_{i\text{d}}\in\mathcal{C}\left(  \left[  0,T_{m}\right]  ,\left[
t_{0},t_{0}+T\right]  \right)  ,i=2,3,4}{\min} &
\begin{array}
[c]{l}%
w_{g}\left\Vert \int_{t_{0}}^{t_{0}+T}\left(  \mathbf{f}_{g}\left(  t\right)
-m\mathbf{g}\right)  \text{d}t\right\Vert ^{2}+w_{t}\left\Vert \int_{t_{0}%
}^{t_{0}+T}\left(  \mathbf{f}_{t}\left(  t\right)  -m\mathbf{a}_{d}\left(
t\right)  \right)  \text{d}t\right\Vert ^{2}\\
+w_{e}\int_{t_{0}}^{t_{0}+T}\left\Vert -\mathbf{f}_{g}\left(  t\right)
+\mathbf{f}_{t}\left(  t\right)  \right\Vert ^{2}\text{d}t
\end{array}
\\
\text{s.t.} &
\begin{array}
[c]{l}%
f_{d}\mathbf{R}\left(  \mathbf{q}_{d}\right)  \mathbf{e}_{3}=-\mathbf{f}%
_{g}+\mathbf{f}_{t}\\
\mathbf{\dot{q}}_{d}=\left[
\begin{array}
[c]{cc}%
0 & -\boldsymbol{\omega}_{d}^{\text{T}}\\
\boldsymbol{\omega}_{d} & -\left[  \boldsymbol{\omega}_{d}\right]  _{\times}%
\end{array}
\right]  \mathbf{q}_{d}\\
\mathbf{J}\boldsymbol{\dot{\omega}}_{d}=-\boldsymbol{\omega}_{d}\times\left(
\mathbf{J}\boldsymbol{\omega}_{d}\right)  +\mathbf{G}_{a}+\boldsymbol{\tau
}_{d}\\
\left[
\begin{array}
[c]{c}%
f_{d}\\
\boldsymbol{\tau}_{d}%
\end{array}
\right]  =\mathbf{HT}_{d},T_{1d}=0.
\end{array}
.
\end{array}
\end{equation}

\subsection{Application to a Fixed-Wing UAV Control}

\begin{figure}[h]
\centering \includegraphics[scale= 0.6]{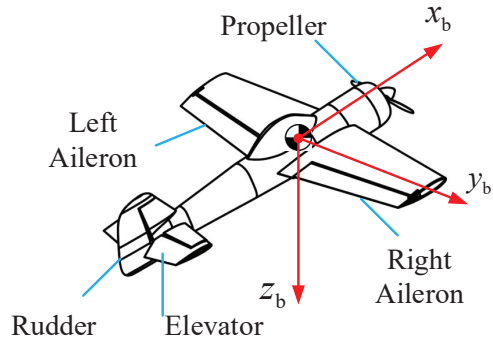}\caption{A fixed-wing
UAV}%
\end{figure}

As shown in Fig.5, a fixed-wing UAV's dynamic model is expressed as
\cite{Zhou(2017)}%
\begin{align}
\mathbf{\dot{p}}  &  =\mathbf{v}\nonumber\\
\mathbf{\dot{v}}  &  =\mathbf{g-}\frac{1}{m}\mathbf{R}\left(  \mathbf{q}%
\right)  \left(  \mathbf{f}_{p}+\mathbf{R}_{a}\left(  \alpha,\beta\right)
\mathbf{f}_{a}\right) \\
\mathbf{\dot{q}}  &  =\left[
\begin{array}
[c]{cc}%
0 & -\boldsymbol{\omega}^{\text{T}}\\
\boldsymbol{\omega} & -\left[  \boldsymbol{\omega}\right]  _{\times}%
\end{array}
\right]  \mathbf{q}\\
\mathbf{J}\boldsymbol{\dot{\omega}}  &  =-\boldsymbol{\omega}\times
\mathbf{J}\boldsymbol{\omega}+\mathbf{G}+\mathbf{m}_{p}+\mathbf{m}_{a}%
\end{align}
where $\mathbf{f}_{p}\in\mathbf{%
\mathbb{R}
}^{3}$ and $\mathbf{m}_{p}\in\mathbf{%
\mathbb{R}
}^{3}$ are the force and torque produced by the propeller; $\mathbf{f}%
_{a},\mathbf{m}_{a}$ are the force and torque produced by the aerodynamic
force; the other physical meaning is the same to those in (\ref{quadcopter})
because both a quadcopter and a fixed-wing UAV are rigid body. Let
$\mathbf{v}_{w}\in\mathbf{%
\mathbb{R}
}^{3}$\ be wind velocity. Define the airspeed $\mathbf{v}_{a}\in\mathbf{%
\mathbb{R}
}^{3}$ as%
\[
\mathbf{v}_{a}=\mathbf{R}^{\text{T}}\left(  \mathbf{v-v}_{w}\right)  .
\]
The airspeed is $V_{a}=\left\Vert \mathbf{v}_{a}\right\Vert $, angle of attack
$\alpha$ and sideslip angle $\beta$ are%
\begin{align*}
\alpha &  =\tan^{-1}\left(  \frac{\mathbf{v}_{a,3}}{\mathbf{v}_{a,1}}\right)
\\
\beta &  =\sin^{-1}\left(  \frac{\mathbf{v}_{a,2}}{\left\Vert \mathbf{v}%
_{a}\right\Vert }\right)
\end{align*}
where $\mathbf{x}_{i}$ is the $i$th element of $\mathbf{x.}$ Let
$\mathbf{v}_{e}\in\mathbf{%
\mathbb{R}
}^{3}$\ be the downwash velocity of the propeller as shown in Fig.6, defined
as%
\[
\mathbf{v}_{e}=\left[
\begin{array}
[c]{c}%
V_{e}\\
0\\
0
\end{array}
\right]
\]
where $V_{e}=k_{m}\delta_{t}$ is the speed of the air as it leaves the
propeller, $k_{m}>0$ and $\delta_{t}\in\left[  0,1\right]  $ is the
pluse-width-modulation command.\begin{figure}[h]
\centering \includegraphics[scale= 0.8]{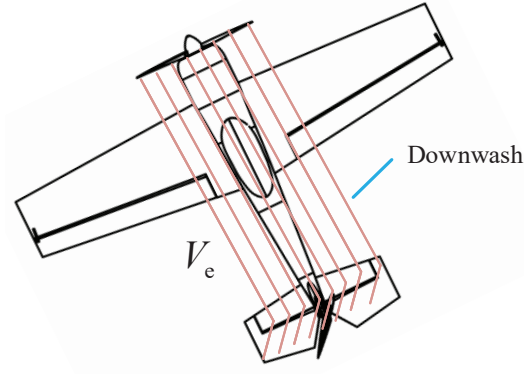}\caption{Downwash}%
\end{figure}

For simplicity, as shown in Fig.6, we assume that the elevator, rudder are
affected \emph{fully} by the downwash \cite{Matsumoto(2010)}. Aerodynamic
force consists of the lift, drag and side force. The lift is perpendicular to
the airspeed while the drag is in line with and opposing the airspeed. The
directions of drag, side force, and lift form a right-handed frame. Similar to
book \cite{Beard(2012)}, we have
\begin{align*}
\mathbf{f}_{p}  &  =\frac{1}{2}\rho S_{p}C_{p}\left[
\begin{array}
[c]{c}%
V_{e}^{2}-V_{a}^{2}\\
0\\
0
\end{array}
\right] \\
\mathbf{m}_{p}  &  =\left[
\begin{array}
[c]{c}%
-k_{T_{p}}\left(  k_{\Omega}\delta_{t}\right)  ^{2}\\
0\\
0
\end{array}
\right]
\end{align*}

\[
\mathbf{f}_{a}=\left[
\begin{array}
[c]{c}%
-\frac{1}{2}\rho V_{a}^{2}SC_{Df}\left(  \alpha,q\right)  -\frac{1}{2}\rho
V_{a}^{2}SC_{Dw_{l}}\left(  \alpha,q,\delta_{al}\right)  -\frac{1}{2}\rho
V_{a}^{2}SC_{Dw_{r}}\left(  \alpha,q,\delta_{ar}\right)  -\frac{1}{2}\rho
V_{e}^{2}SC_{D_{e}}\left(  \delta_{e}\right)  \\
\frac{1}{2}\rho V_{a}^{2}SbC_{lf}\left(  \beta,p,r\right)  +\frac{1}{2}\rho
V_{a}^{2}SbC_{lw_{l}}\left(  \beta,p,r,\delta_{al}\right)  +\frac{1}{2}\rho
V_{a}^{2}SbC_{lw_{r}}\left(  \beta,p,r,\delta_{ar}\right)  +\frac{1}{2}\rho
V_{e}^{2}SbC_{l_{r}}\left(  \delta_{r}\right)  \\
\frac{1}{2}\rho V_{a}^{2}SC_{Lf}\left(  \alpha,q\right)  +\frac{1}{2}\rho
V_{a}^{2}SC_{Lw_{l}}\left(  \alpha,q,\delta_{al}\right)  +\frac{1}{2}\rho
V_{a}^{2}SC_{Lw_{r}}\left(  \alpha,q,\delta_{ar}\right)  +\frac{1}{2}\rho
V_{e}^{2}SC_{Le}\left(  \delta_{e}\right)
\end{array}
\right]
\]

\[
\mathbf{m}_{a}=\left[
\begin{array}
[c]{c}%
l\\
m\\
n
\end{array}
\right]  =\left[
\begin{array}
[c]{c}%
\frac{1}{2}\rho V_{a}^{2}SbC_{lf}\left(  \beta,p,r\right)  +\frac{1}{2}\rho
V_{a}^{2}SbC_{lw_{l}}\left(  \beta,p,r,\delta_{al}\right)  \\
\text{ \ \ \ \ \ \ \ \ \ }+\frac{1}{2}\rho V_{a}^{2}SbC_{lw_{r}}\left(
\beta,p,r,\delta_{ar}\right)  +\frac{1}{2}\rho V_{e}^{2}SbC_{l_{r}}\left(
\delta_{r}\right)  \\
\frac{1}{2}\rho V_{a}^{2}ScC_{mf}\left(  \alpha,q\right)  +\frac{1}{2}\rho
V_{a}^{2}ScC_{mw_{l}}\left(  \alpha,q,\delta_{al}\right)  \\
\text{ \ \ \ \ \ \ \ \ \ }+\frac{1}{2}\rho V_{a}^{2}ScC_{mw_{r}}\left(
\alpha,q,\delta_{ar}\right)  +\frac{1}{2}\rho V_{w}^{2}ScC_{m_{e}}\left(
\delta_{e}\right)  \\
\frac{1}{2}\rho V_{a}^{2}SbC_{nf}\left(  \beta,p,r\right)  +\frac{1}{2}\rho
V_{a}^{2}SbC_{nw_{l}}\left(  \beta,p,r,\delta_{al}\right)  \\
\text{ \ \ \ \ \ \ \ \ \ }+\frac{1}{2}\rho V_{a}^{2}SbC_{nw_{r}}\left(
\beta,p,r,\delta_{ar}\right)  +\frac{1}{2}\rho V_{w}^{2}SbC_{n_{r}}\left(
\delta_{r}\right)
\end{array}
\right]
\]
where $\rho$ is the air density, $S$ is the wing area, $b$ is the wingspan,
$c$ is the mean chord of the wing, $S_{p}$ is the area swept out by the
propeller, $k_{\Omega}\delta_{t}$ is the propeller speed and $k_{T_{p}}$ is a
constant determined by experiment, $C_{p}$ is a constant determined by
experiment; the aerodynamic force and moments are produced by the \emph{left
wing including its aileron}, \emph{the right wing including its aileron},
\emph{the fuselage}, \emph{elevator} and \emph{rudder}; the left and right
aileron angles are denoted by $\delta_{al},\delta_{ar}\in\lbrack-\delta
_{am,}\delta_{am,}]$, elevator angle by $\delta_{e}\in\lbrack-\delta
_{em,}\delta_{em,}]$ and rudder angle by $\delta_{r}\in\lbrack-\delta
_{rm,}\delta_{rm,}].$ Furthermore, the aerodynamic force and moments are
produced by both the airflow and downwash, where the physical meaning of
components is shown in Tables 1-2. Due to the symmetry of the left and right
ailerons, we have
\begin{align*}
C_{Dw_{l}}\left(  \alpha,q,\cdot\right)   &  =-C_{Dw_{r}}\left(
\alpha,q,\cdot\right)  \\
C_{lw_{l}}\left(  \beta,p,r,\cdot\right)   &  =-C_{lw_{r}}\left(
\beta,p,r,\cdot\right)  \\
C_{Lw_{l}}\left(  \alpha,q,\cdot\right)   &  =-C_{Lw_{r}}\left(
\alpha,q,\cdot\right)  \\
C_{lw_{l}}\left(  \beta,p,r,\cdot\right)   &  =-C_{lw_{r}}\left(
\beta,p,r,\cdot\right)  \\
C_{mw_{l}}\left(  \alpha,q,\cdot\right)   &  =-C_{mw_{r}}\left(
\alpha,q,\cdot\right)  \\
C_{nw_{l}}\left(  \beta,p,r,\cdot\right)   &  =-C_{nw_{r}}\left(
\beta,p,r,\cdot\right)  .
\end{align*}

\bigskip

\begin{center}
Table 1. Aerodynamic force components
\end{center}%

\[%
\begin{tabular}
[c]{|l|l|}\hline
Symbol & Meaning\\\hline
$-\frac{1}{2}\rho V_{a}^{2}SC_{Df}\left(  \alpha,q\right)  $ & Drag by the
fuselage\\\hline
$-\frac{1}{2}\rho V_{a}^{2}SC_{Dw_{l}}\left(  \alpha,q,\delta_{al}\right)  $ &
Drag by the left wing including its aileron\\\hline
$-\frac{1}{2}\rho V_{a}^{2}SC_{Dw_{r}}\left(  \alpha,q,\delta_{ar}\right)  $ &
Drag by the right wing including its aileron\\\hline
$\frac{1}{2}\rho V_{e}^{2}SC_{D_{e}}\left(  \delta_{e}\right)  $ & Drag by the
rudder caused by downwash\\\hline
$\frac{1}{2}\rho V_{a}^{2}SbC_{lf}\left(  \beta,p,r\right)  $ & Side force by
the fuselage\\\hline
$\frac{1}{2}\rho V_{a}^{2}SbC_{lw_{l}}\left(  \beta,p,r,\delta_{al}\right)  $
& Side force by the left wing including its aileron\\\hline
$\frac{1}{2}\rho V_{a}^{2}SbC_{lw_{r}}\left(  \beta,p,r,\delta_{ar}\right)  $
& Side force by the right wing including its aileron\\\hline
$\frac{1}{2}\rho V_{e}^{2}SbC_{l_{r}}\left(  \delta_{r}\right)  $ & Side force
by the rudder caused by downwash\\\hline
$\frac{1}{2}\rho V_{a}^{2}SC_{Lf}\left(  \alpha,q\right)  $ & Lift by the
fuselage\\\hline
$\frac{1}{2}\rho V_{a}^{2}SC_{Lw_{l}}\left(  \alpha,q,\delta_{al}\right)  $ &
Lift by the left wing including its aileron\\\hline
$\frac{1}{2}\rho V_{a}^{2}SC_{Lw_{r}}\left(  \alpha,q,\delta_{ar}\right)  $ &
Lift by the right wing including its aileron\\\hline
$\frac{1}{2}\rho V_{e}^{2}SC_{Le}\left(  \delta_{e}\right)  $ & Lift by the
rudder caused by downwash\\\hline
\end{tabular}
\]

\bigskip

\bigskip\bigskip\bigskip\bigskip\bigskip\bigskip

\bigskip

\begin{center}
Table 2. Aerodynamic moment components
\end{center}%

\[%
\begin{tabular}
[c]{|l|l|}\hline
Symbol & Meaning\\\hline
$\frac{1}{2}\rho V_{a}^{2}SbC_{lf}\left(  \beta,p,r\right)  $ & Roll moment by
the fuselage\\\hline
$\frac{1}{2}\rho V_{a}^{2}SbC_{lw_{l}}\left(  \beta,p,r,\delta_{al}\right)  $
& Roll moment by the left wing including its aileron\\\hline
$\frac{1}{2}\rho V_{a}^{2}SbC_{lw_{r}}\left(  \beta,p,r,\delta_{ar}\right)  $
& Roll moment by the right wing including its aileron\\\hline
$\frac{1}{2}\rho V_{e}^{2}SbC_{l_{r}}\left(  \delta_{r}\right)  $ & Roll
moment by the rudder caused by downwash\\\hline
$\frac{1}{2}\rho V_{a}^{2}ScC_{mf}\left(  \alpha,q\right)  $ & Pitch moment by
the fuselage\\\hline
$\frac{1}{2}\rho V_{a}^{2}ScC_{mw_{l}}\left(  \alpha,q,\delta_{al}\right)  $ &
Pitch moment by the left wing including its aileron\\\hline
$\frac{1}{2}\rho V_{a}^{2}ScC_{mw_{r}}\left(  \alpha,q,\delta_{ar}\right)  $ &
Pitch moment by the right wing including its aileron\\\hline
$\frac{1}{2}\rho V_{w}^{2}ScC_{m_{e}}\left(  \delta_{e}\right)  $ & Pitch
moment by the elevator\\\hline
$\frac{1}{2}\rho V_{a}^{2}SbC_{nf}\left(  \beta,p,r\right)  $ & Yaw moment by
the fuselage\\\hline
$\frac{1}{2}\rho V_{a}^{2}SbC_{nw_{l}}\left(  \beta,p,r,\delta_{al}\right)  $
& Yaw moment by the left wing including its aileron\\\hline
$\frac{1}{2}\rho V_{a}^{2}SbC_{nw_{r}}\left(  \beta,p,r,\delta_{ar}\right)  $
& Yaw moment by the right wing including its aileron\\\hline
$\frac{1}{2}\rho V_{w}^{2}SbC_{n_{r}}\left(  \delta_{r}\right)  $ & Yaw moment
by the rudder caused by downwash\\\hline
\end{tabular}
\ \
\]

The objective is to design $\delta_{t},\delta_{al},\delta_{ar},\delta
_{e},\delta_{r}$ to replicate the accident shown in Fig.1. First, make the UAV
track $\mathbf{p}_{d}\left(  t\right)  $\ as accurately as possible when no
failure happens. Then, make it land safely when the left half of the wings
loses completely. By using (\ref{desiredacc}), we hope the desired total
propeller thrust and desired rotation matrix satisfy%
\[
\mathbf{g-}\frac{1}{m}\mathbf{R}\left(  \mathbf{q}_{d}\right)  \left(
\mathbf{f}_{pd}+\mathbf{R}_{a}\left(  \alpha_{d},\beta_{d}\right)
\mathbf{f}_{ad}\right)  =\mathbf{a}_{d}%
\]
where $\mathbf{f}_{pd},\mathbf{q}_{d},\alpha_{d},\beta_{d},\mathbf{f}_{ad}$
are finally determined by $\delta_{t},\delta_{al},\delta_{ar},\delta
_{e},\delta_{r}.$ If the left half of wings loses completely, then
$\delta_{al}=0$ and $C_{\cdot w_{l}}=0.$ Since
\begin{equation}
\mathbf{f}_{d}=\mathbf{-}\frac{1}{m}\mathbf{R}\left(  \mathbf{q}_{d}\right)
\left(  \mathbf{f}_{pd}+\mathbf{R}_{a}\left(  \alpha_{d},\beta_{d}\right)
\mathbf{f}_{ad}\right)
\end{equation}
we can obtain that%
\begin{equation}%
\begin{array}
[c]{ll}%
\underset{}{\underset{\delta_{ed}\in\lbrack-\delta_{em,}\delta_{em,}%
],\delta_{rd}\in\lbrack-\delta_{rm,}\delta_{rm,}]}{\underset{\delta_{td}%
\in\left[  0,1\right]  ,\delta_{ald},\delta_{ard}\in\lbrack-\delta_{am,}%
\delta_{am,}],}{\min}}} &
\begin{array}
[c]{l}%
w_{g}\left\Vert \int_{t_{0}}^{t_{0}+T}\left(  \mathbf{f}_{g}\left(  t\right)
-m\mathbf{g}\right)  \text{d}t\right\Vert ^{2}+w_{t}\left\Vert \int_{t_{0}%
}^{t_{0}+T}\left(  \mathbf{f}_{t}\left(  t\right)  -m\mathbf{a}_{d}\left(
t\right)  \right)  \text{d}t\right\Vert ^{2}\\
+w_{e}\int_{t_{0}}^{t_{0}+T}\left\Vert -\mathbf{f}_{g}\left(  t\right)
+\mathbf{f}_{t}\left(  t\right)  \right\Vert ^{2}\text{d}t
\end{array}
\\
\text{s.t.} &
\begin{array}
[c]{l}%
\mathbf{-}\frac{1}{m}\mathbf{R}\left(  \mathbf{q}_{d}\right)  \left(
\mathbf{f}_{pd}\left(  \delta_{td}\right)  +\mathbf{R}_{a}\left(  \alpha
_{d},\beta_{d}\right)  \mathbf{f}_{ad}\left(  \delta_{td},\delta_{ald}%
,\delta_{ard},\delta_{ed},\delta_{rd}\right)  \right) \\
=-\mathbf{f}_{g}+\mathbf{f}_{t}\\
\mathbf{\dot{q}}_{d}=\left[
\begin{array}
[c]{cc}%
0 & -\boldsymbol{\omega}_{d}^{\text{T}}\\
\boldsymbol{\omega}_{d} & -\left[  \boldsymbol{\omega}_{d}\right]  _{\times}%
\end{array}
\right]  \mathbf{q}_{d}\\
\mathbf{J}\boldsymbol{\dot{\omega}}_{d}=-\boldsymbol{\omega}_{d}%
\times\mathbf{J}\boldsymbol{\omega}_{d}+\mathbf{G}+\mathbf{m}_{p}\left(
\delta_{td}\right)  +\mathbf{m}_{a}\left(  \delta_{td},\delta_{ald}%
,\delta_{ard},\delta_{ed},\delta_{rd}\right)
\end{array}
\end{array}
\end{equation}
according to the gravity-compensation-first principle from the impulse
viewpoint. Furthermore, if the left half of wings loses completely, then%
\begin{equation}%
\begin{array}
[c]{ll}%
\underset{\delta_{ed}\in\lbrack-\delta_{em,}\delta_{em,}],\delta_{rd}%
\in\lbrack-\delta_{rm,}\delta_{rm,}]}{\underset{\delta_{td}\in\left[
0,1\right]  ,\delta_{ard}\in\lbrack-\delta_{am,}\delta_{am,}],}{\min}} &
\begin{array}
[c]{l}%
w_{g}\left\Vert \int_{t_{0}}^{t_{0}+T}\left(  \mathbf{f}_{g}\left(  t\right)
-m\mathbf{g}\right)  \text{d}t\right\Vert ^{2}+w_{t}\left\Vert \int_{t_{0}%
}^{t_{0}+T}\left(  \mathbf{f}_{t}\left(  t\right)  -m\mathbf{a}_{d}\left(
t\right)  \right)  \text{d}t\right\Vert ^{2}\\
+w_{e}\int_{t_{0}}^{t_{0}+T}\left\Vert -\mathbf{f}_{g}\left(  t\right)
+\mathbf{f}_{t}\left(  t\right)  \right\Vert ^{2}\text{d}t
\end{array}
\\
\text{s.t.} &
\begin{array}
[c]{l}%
\mathbf{-}\frac{1}{m}\mathbf{R}\left(  \mathbf{q}_{d}\right)  \left(
\mathbf{f}_{pd}\left(  \delta_{td}\right)  +\mathbf{R}_{a}\left(  \alpha
_{d},\beta_{d}\right)  \mathbf{f}_{ad}\left(  \delta_{td},\delta_{ald}%
,\delta_{ard},\delta_{ed},\delta_{rd}\right)  \right) \\
=-\mathbf{f}_{g}+\mathbf{f}_{t}\\
\mathbf{\dot{q}}_{d}=\left[
\begin{array}
[c]{cc}%
0 & -\boldsymbol{\omega}_{d}^{\text{T}}\\
\boldsymbol{\omega}_{d} & -\left[  \boldsymbol{\omega}_{d}\right]  _{\times}%
\end{array}
\right]  \mathbf{q}_{d}\\
\mathbf{J}\boldsymbol{\dot{\omega}}_{d}=-\boldsymbol{\omega}_{d}\times\left(
\mathbf{J}\boldsymbol{\omega}_{d}\right)  +\mathbf{G}+\mathbf{m}_{p}\left(
\delta_{td}\right)  +\mathbf{m}_{a}\left(  \delta_{td},\delta_{ald}%
,\delta_{ard},\delta_{ed},\delta_{rd}\right) \\
\delta_{ald}=0,\text{ }C_{\cdot w_{l}}=0.
\end{array}
.
\end{array}
\end{equation}

\section{CONCLUSIONS}

According to the first principle, the gravity-compensation-first principle is
developed. It states that all control gives priority to compensating for the
effect caused by the gravity of the aircraft, with the left authority used to
meet the requirements of the desired motion of the aircraft. According to
Newton's second law, the principle is further formulated as optimization
problems in three aspects: force, impulse, and energy based on a mass point
model. Furthermore, the principle is applied to a quadcopter and a fixed-wing
UAV with severe failure. Model predictive control methods can help solve the
problems by the gravity-compensation-first principle.

\end{document}